\documentclass[10pt,twocolumn,letterpaper]{article}

\usepackage[pagenumbers]{cvpr} % To force page numbers, e.g. for an arXiv version

\usepackage{graphicx}
\usepackage{amsmath}
\usepackage{amssymb}
\usepackage{booktabs}

\usepackage{float}
\usepackage{tabularx}

\usepackage[pagebackref,breaklinks,colorlinks]{hyperref}

\usepackage[capitalize]{cleveref}
\crefname{section}{Sec.}{Secs.}
\Crefname{section}{Section}{Sections}
\Crefname{table}{Table}{Tables}
\crefname{table}{Tab.}{Tabs.}

\def\cvprPaperID{16} % *** Enter the CVPR Paper ID here
\def\confName{CVPR}
\def\confYear{2022}

\begin{document}

% \title{\LaTeX\ Author Guidelines for \confName~Proceedings}
\title{LayoutBERT: Masked Language Layout Model for Object Insertion}

\author{Kerem Turgutlu\\
Adobe\\
{\tt\small turgutlu@adobe.com}
\and
Sanat Sharma\\
Adobe\\
{\tt\small sanatsha@adobe.com}
\and
Jayant Kumar\\
Adobe\\
{\tt\small jaykumar@adobe.com}
}
\maketitle

%%%%%%%%% ABSTRACT
\begin{abstract}
Image compositing is one of the most fundamental steps in creative workflows. It involves taking objects/parts of several images to create a new image, called a composite. Currently, this process is done manually by creating accurate masks of objects to be inserted and carefully blending them with the target scene or images, usually with the help of tools such as Photoshop or GIMP. While there have been several works on automatic selection of objects for creating masks, the problem of object placement within an image with the correct position, scale, and harmony remains  a difficult problem with limited exploration. Automatic object insertion in images or designs is a difficult problem as it requires understanding of the scene geometry and the color harmony between objects. 

We propose \textbf{LayoutBERT} for the object insertion task. It uses a novel self-supervised masked language model objective and bidirectional multi-head self-attention. It outperforms previous layout-based likelihood models and shows favorable properties in terms of model capacity. We demonstrate the effectiveness of our approach for object insertion in the image compositing setting and other settings like documents and design templates. We further demonstrate the usefulness of the learned representations for layout-based retrieval tasks. We provide both qualitative and quantitative evaluations on datasets from diverse domains like COCO, PublayNet, and two  new datasets which we  call Image Layouts and Template Layouts. Image Layouts which consists of 5.8 million images with layout annotations is the largest image layout dataset to our knowledge. We also share ablation study results on the effect of dataset size, model size and class sample size for this task.

\end{abstract}

%%%%%%%%% BODY TEXT
\section{Introduction}
\label{sec:intro}
With the recent rise in image and video creation for teaching, advertisement, information sharing, and social 
%%% don't touch
\begin{figure}[htp!]
    \begin{center}
    \includegraphics[width=\linewidth]{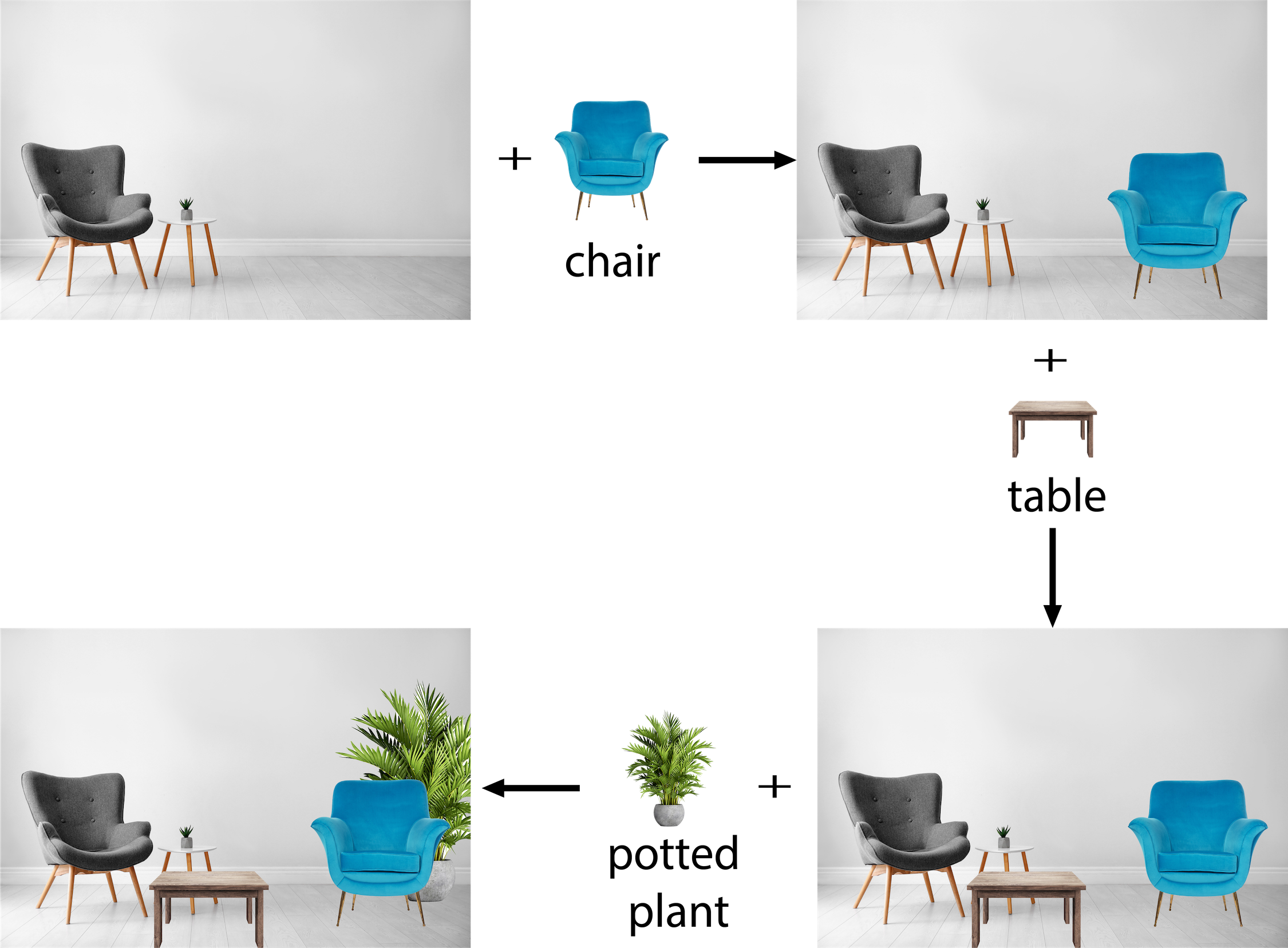}
    \end{center}
    \caption{Iterative class conditional compositing using LayoutBert bounding box predictions and alpha masking. At each step, object insertion orders are re-sorted based on bottom bounding box coordinates to avoid unrealistic occlusion.}
    \label{fig:teaser}
\end{figure}
%%%
influencing, the need for AI-based assistance in image and video editing is greater than ever. Image Compositing is one of the most common tasks in creative workflows. However, currently, it involves several manual steps such as background and foreground selection, masking, refinement, placement, scale-adjustment, and harmonization. Due to this tedious and multi-step process it is difficult for creatives to try more than few new design ideas. Moreover, the learning curve for beginners is quite steep making it inaccessible for the majority of users who are interested in expressing themselves in a creative way.

The success of self-attention and transformer networks on several key language and vision learning tasks has led to several new  explorations of these architectures including layout understanding and generation. 
In this work, we further push the state-of-the-art on layout understanding and show very promising results on realistic images as well as on documents and templates. We envision a future where layout input from the user is the exception rather than the norm.

\begin{figure*}
    \begin{center}
    \includegraphics[width=\linewidth,scale=2]{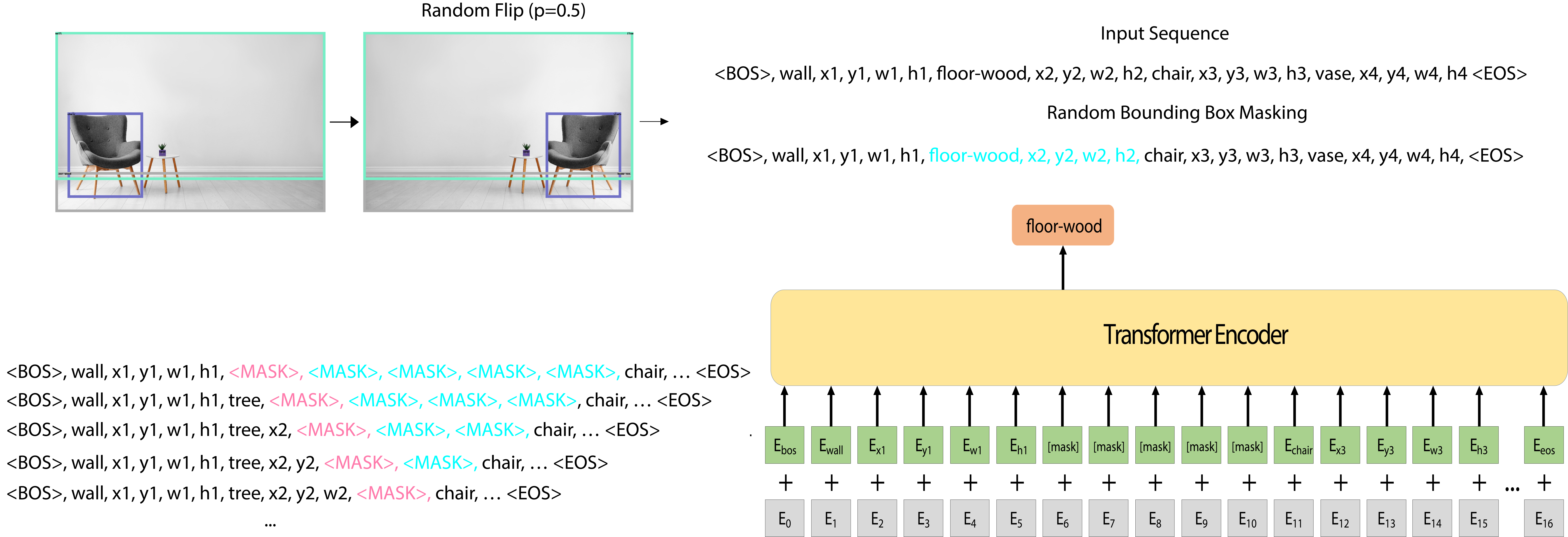}
    \end{center}
    \caption{Masked input sequences during LayoutBert training. Random left-right flip is applied during training as data augmentation.  Later the 2D layout is converted into the input sequence for modeling. A bounding box is selected for masking with uniform sampling and each token of the selected bounding box is masked iteratively and added to the batch. For each added sample, the model predicts the left-most masked token (denoted by pink coloring).}
    \label{fig:main_diagram}
\end{figure*}

Usually, creators/designers start with a blank canvas, initialize their work with a base or background image and bring in parts from multiple images while applying geometric and color transformations (edits) until the desired creation is achieved. The final creation can be a personal family collage, an advertisement photo, a sci-fi movie poster, a petting zoo fundraiser flyer, etc.

There are many existing works based on automated color transformation learning such as deep image harmonization \cite{tsai2017deep, ling2021region}. However, none of them discusses geometric transformation learning for image compositing or the existing work often limits the problem to certain classes or less complex datasets.

In this paper we propose a bidirectional likelihood based model which can learn the most likely location and scale for a given object to be inserted into an image. Our approach offers a solution at scale to automate both photo-realistic and template-like object insertion conditioned on the desired class using BERT \cite{devlin2018bert} with a custom self-supervised optimization objective. We also provide layout based retrieval results using the representations learned from the self-supervised training. Our main contributions are:

\begin{itemize}
\item A novel self-supervised masked language model LayoutBERT for layout understanding,
\item Application to object insertion and layout retrieval using LayoutBERT,
\item State-of-the-art results on multiple datasets and ablation studies on the largest known image layout dataset.
\end{itemize}

\section{Related Work}
\label{sec:formatting}

There are several studies in the literature on deep image compositing \cite{lin2018st,lee2018context} which use generative adversarial networks (GANs) \cite{goodfellow2020generative} and specialized networks such as spatial transformer networks (STNs) \cite{jaderberg2015spatial}. \cite{lin2018st} trains sequential STN generators: It iteratively applies geometric transformations (image warping) on an initially placed foreground image and tries to generate a realistic final composite image. \cite{lee2018context} proposes two network modules, (1) \emph{where} and (2) \emph{what} modules. The \emph{where} module generates realistic bounding boxes by transforming a unit bounding box with a STN and the \emph{what} module uses that bounding box to generate a semantic map for the desired class instance while optimizing to make the final semantic map realistic. A separate model is trained for each class, \eg a pedestrian model and a car model in the case of the Cityscapes dataset \cite{cordts2016cityscapes}. Unfortunately, neither of these approaches can be scaled to larger model and to many classes.

Although GANs are proven to work very well on realistic image generation with high fidelity, they are notoriously difficult to train and are not easily scalable. Self-attention and transformer networks have been successfully used to train billion parameter models and even a trillion parameters language model \cite{roberts2020much}. Recently, multi-head self-attention has been used to train models for layout generation and completion \cite{gupta2021layouttransformer, arroyo2021variational}. An image can be represented as a set of layout elements by extracting class and bounding box information of the overall scene and of the objects in it. Similarly, a text document can be represented by the bounding box information of different elements such as titles, texts, graphs, and figures. The same idea can be extended to any creative design like posters, flyers, invitation cards, etc. In this regard, \cite{gupta2021layouttransformer} trains an autoregressive model using a causal attention mask and maximizes log-likelihood via a next token prediction task similar to \cite{radford2018improving} in a self-supervised manner. \cite{arroyo2021variational} extends this idea by proposing a variational autoencoder (VAE) to learn better representations using a BERT \cite{devlin2018bert} like encoder and a GPT \cite{radford2018improving} like decoder. Layout transformer models use only the class and bounding box information extracted from the image pixel data. This can be considered as a disadvantage over CNN-based models in terms of the granularity of information, but this same property allows more efficient training and inference, hence scalability.

Although previous layout transformer models can model the data likelihood and the distribution, they are not optimal for the object insertion task. \cite{gupta2021layouttransformer} can generate or complete layouts but it can only attend to left context and cannot see the whole scene at once, which is a major drawback for the object insertion task. \cite{arroyo2021variational} can learn better representations and improve generation diversity but that usually comes with the cost of likelihood. \cite{lee2018context} directly learns where to generate bounding boxes for object insertion with its \emph{where} module; however it requires training of a separate model for every class. Also, a GAN objective makes it difficult to use 'too powerful' discriminators or generators. We argue that the object insertion task requires seeing the whole scene at once, modeling long range dependencies in the scene with a module like self-attention and large scale models. Our work combines and extends ideas from prior art and can be considered similar to the \emph{where} module of \cite{lee2018context} in terms of the task and to \cite{gupta2021layouttransformer} in terms of the input representation and multi-head self-attention mechanism. Layout understanding is similar to scene understanding and using bidirectional context is a natural choice for learning realistic layouts.

%------------------------------------------------------------------------
%%%%% Layout Bert
\section{LayoutBERT}

Our method treats image, document or template layouts as scene graphs and tries to solve the object insertion task using a masked language modeling objective. For this purpose we use a bidirectional transformer model like BERT due to its popularity and name our method accordingly. That been said, our custom masked language modeling objective for layout understanding can be used with any of the transformer models like \cite{kitaev2020reformer,wang2020linformer,choromanski2020rethinking,beltagy2020longformer} and even with a bidirectional LSTM, GRU, or RNN. Original BERT model for NLP language modeling is trained using two tasks: masked LM and next sentence prediction (NSP). However, neither of these tasks are suitable for layout understanding and object insertion. Masked LM objective picks individual tokens to be masked during training, in contrast we iteratively mask a span of tokens which correspond to a bounding box. Next Sentence Prediction is used for classifying whether a sentence comes after another given sentence and cannot be used for token generation in the context of the object insertion task. Also, \cite{liu2019roberta} shows that removing the NSP loss matches or slightly improves downstream NLP task performance, hence not critical.

\subsection{Layout Representation}

To create input sequences for training we primarily follow the ideas from \cite{gupta2021layouttransformer}. We use bounding box annotations of raw images, documents or templates depending on the dataset. When bounding box annotations are not available and during inference time we use a pretrained panoptic segmentation model \cite{kirillov2019panoptic} in the case of images and an object detection model in the case of documents and templates. We convert bounding boxes into a flat sequence using the raster scan order. A sequence input is then represented as: \emph{BOS, $c_{1}, x_{1}, y_{1}, w_{1}, h_{1}, c_{2}, x_{2}, y_{2}, w_{2}, h_{2}$, …, EOS}, where \emph{c,x,y,w,h} stands for class token, top-left x coordinate, top-left y coordinate, width and height tokens respectively. \emph{BOS} and \emph{EOS} are special tokens: beginning and end of sentence. During tokenization each class id is mapped to a unique class token, and \emph{x,y,w,h} tokens are mapped to discrete space by splitting the 2D input into a $NXN$ grid similar to \emph{anchors} from the object detection literature.

\subsection{Model Architecture and Training}

We use BERT architecture introduced in \cite{devlin2018bert} and optimize it using a novel self-supervised training objective for layout understanding. For the object insertion task it is important to look at the whole context at once for generating bounding boxes. This is where bidirectional attention comes to play. During training we randomly select a bounding box and mask the 5 tokens \emph{c,x,y,w,h} which represent it. We do it by creating 5 duplicates for each sequence sample and masking all 5 tokens in the beginning and iteratively unmasking the left-most token at each step as shown in \ref{fig:main_diagram}. For each masked sequence we try to predict the left-most masked token.
This custom masked language modeling objective allows our model to generate bounding boxes by predicting c,x,y,w and h in a step-by-step fashion by mimicking autoregressive likelihood models in the context of a single bounding box generation while being able to attend to all the other bounding boxes with bidirectional attention. During training time we also apply random left-right flip as data augmentation on the 2D layout before converting it into a flat input sequence.

Instead of using teacher forcing with a decoder-only model like \cite{gupta2021layouttransformer}, we use a BERT architecture with bidirectional attention and model the joint distribution as:

%%% FIX IT
\begin{equation}
\begin{aligned}[b]
    p(\theta_{1:n_{i}}) = \prod_{j=1}^{n_{i}} p(\theta_{j} |\theta_{1: j-1}, \theta_{j + 5 - i + 1:n_{i}})
\end{aligned}
\label{joint_probability}
\end{equation}

\noindent where $n_i = 5(n-1)+i$ is the $i^{th}$ element of the $n^{th}$ box. For example, $i=1$ for $c$, $i=2$ for $x$, $i=3$ for $y$, $i=4$ for $w$, and $i=5$ for $h$.

%------------------------------------------------------------------------
\section{Experiments}

In our evaluations, we refer to \cite{gupta2021layouttransformer} as LayoutTransformer, our re-implementation of \cite{gupta2021layouttransformer} using GPT \cite{radford2018improving} model as LayoutGPT and all other methods with their standard names.

\subsection{Datasets}

We evaluate our results on diverse datasets with natural scenes, documents, and design templates. We use a separate hold-out set for training and validate all models using the official validation splits for each dataset. 

\textbf{COCO} \cite{lin2014microsoft} is a natural scene dataset with common objects which contains both object class and \emph{'stuff'} class annotations. The object class contains a predefined set of 80 common objects while the stuff class contains 92 non-object annotations like sky, wall, grass and pavement. Stuff annotations are as important as object annotations for understanding the overall scene and for the object insertion task. We used COCO Panoptic 2017 dataset and followed the same preparation steps as \cite{gupta2021layouttransformer}. This resulted in 118,280 layouts in the training split and 5,000 layouts in the validation split. The \emph{other} class in the stuff annotations is ignored, resulting in final 80 thing and 91 stuff classes. 

\textbf{PublayNet} \cite{zhong2019publaynet} is a public large scale dataset for document layout understanding. It has 5 categories: text, title, figure, list and table. Similarly, we followed \cite{gupta2021layouttransformer} for data preparation steps including removing layouts with more than 128 elements. This resulted in 335,682 and 11,245 documents layouts for training and validation splits respectively.

\textbf{Image Layouts} is a large scale image dataset with 5.8 million images crawled from the web. Manually annotating such a large dataset is  expensive and labor intensive. For that reason we used a pretrained panoptic segmentation model \cite{kirillov2019panoptic} available at \cite{wu2019detectron2} to generate stuff and object class bounding box annotations. It has total of 133 stuff and object classes.

\textbf{Template Layouts} is a dataset with creative design templates such as posters, flyers, collages, social media posts, ads and more. It has total of 45k templates and 2 classes image and text.

The Image Layouts and Template Layouts datasets are not publicly available, and are curated for experimentation on large scale layout understanding in diverse domains.

\subsection{Object Insertion}\label{object-insertion}

Our custom masked language model objective allows bounding box generation given a layout sequence. This is useful for the object insertion task since our model can predict the most likely class, location and scale to be inserted. LayoutBERT is designed with the object insertion task in mind and can attend to all the bounding boxes in the scene at once. In contrast, previous work \cite{gupta2021layouttransformer} is trained using a decoder-only autoregressive transformer model for generation and can only attend to the left context. During our qualitative analysis we used our own implementation of \cite{gupta2021layouttransformer}, LayoutGPT, since there were no open-source code or models available for conducting our experiments. Our re-implementation outperforms the results reported in the original paper.

\textbf{Object Class Recommendation}. We can make class recommendations about which foreground objects are more likely to be inserted into a given image, document or template. For this purpose, we feed the input sequence to the model and get the output probabilities at each token which give us the most likely predictions for the next token. In the case of the LayoutBERT model, we insert 5 sequential masked tokens in every possible sequence location and get the output probabilities for the masked class token. 

For a simple sequence like \emph{BOS, c, x, y, w, h, EOS} with only 1 bounding box, all possible mask insertions can be seen as:

\begin{itemize}
\item \textbf{Predict at position 1}: \emph{\small{BOS, \textbf{[MASK]}, [MASK], [MASK], [MASK], [MASK], c, x, y, w, h, EOS}}
\item \textbf{Predict at position 2}: \emph{\small{BOS, c, x, y, w, h, \textbf{[MASK]}, [MASK], [MASK], [MASK], [MASK], EOS}}
\end{itemize}

\noindent where we get output probabilities for the masked class tokens denoted in bold. Using these probabilities, we identify the most likely classes that can be inserted after a given partial sequence and later use it for class conditional bounding box generation. We provide top-1 accuracy results for class recommendations in Table \ref{table:top-1-class-rec}. The bidirectional masked language objective allows LayoutBERT to learn correct object classes in all three datasets, showing a significant improvements compared to previous approach.

\begin{table}[H]
    \begin{center}
    \begin{tabular}{c c c c}
    \toprule
     & COCO & PublayNet & Template Lay.  \\
    \midrule
    LayoutGPT & 0.30 & 0.80 & 0.78  \\
    Ours & \textbf{0.44} & \textbf{0.95} & \textbf{0.86}\\
    \bottomrule
    \end{tabular}
    \end{center}
    \caption{Top-1 accuracy class recommendation on COCO, PublayNet and Template Layouts. Higher is better.}
    \label{table:top-1-class-rec}
\end{table}

\textbf{Bounding Box Generation}. When the target class to be inserted is known, either provided or recommended by the model, we can use the model probability outputs at each token to identify the most likely sequence of locations to insert the class token for bounding box generation. For generation with the LayoutGPT model, we use beam search with top-k and top-p sampling with values k=15 and p=0.9. For generation with LayoutBERT model, we use top-k sampling where k=3. These values can be modified to control the level of diversity in bounding box generation. Since LayoutGPT uses a causal mask, it only attends to the previous tokens and is not able to incorporate bidirectional context like LayoutBERT does. To incorporate bidirectional context to LayoutGPT model during generation, we apply left-right flip as a test time augmentation (TTA). For a simple sequence like \emph{BOS, c, x, y, w, h, EOS} with only  1  bounding  box, class conditional iterative bounding box generation at index position 1 can be seen as: 

\begin{itemize}
\item \textbf{Predict x}: \emph{\small{BOS, c, \textbf{[MASK]}, [MASK], [MASK], [MASK], c, x, y, w, h, EOS}}
\item \textbf{Predict y}: \emph{\small{BOS, c, x, \textbf{[MASK]}, [MASK], [MASK], c, x, y, w, h, EOS}}
\item \textbf{Predict w}: \emph{\small{BOS, c, x, y, \textbf{[MASK]}, [MASK], c, x, y, w, h, EOS}}
\item \textbf{Predict h}: \emph{\small{BOS, c, x, y, w, \textbf{[MASK]}, c, x, y, w, h, EOS}}
\end{itemize}

\noindent We provide mean intersection over union (mIoU) results on class conditional bounding box generations in Table \ref{table:miou-box-gen}. It is calculated by taking the average of predicted and ground truth bounding box IoUs for a given sequence.

\begin{table}[H]
    \begin{center}
    \begin{tabular}{c c c c}
    \toprule
     & COCO & PublayNet & Template Lay.  \\
    \midrule
    LayoutGPT & 0.31  & 0.52 & 0.21  \\
    Ours & \textbf{0.36} & \textbf{0.78} & \textbf{0.27} \\
    \bottomrule
    \end{tabular}
    \end{center}
    \caption{Class conditional bounding box generation mIoU on COCO, PublayNet and Template Layouts. Higher is better.}
    \label{table:miou-box-gen}
\end{table}

\textbf{Scoring and Post-Processing}. Each bounding box generation has an associated output probability. This allows us to score any predicted bounding box \emph{x, y, w, h} and use this score for ranking. We apply non-maximum suppression (NMS) on our bounding box generations similar to popular object detection models \cite{girshick2015fast}. This allows us to remove bounding boxes with lower scores that overlap too much. The NMS threshold is another controllable parameter like top-k and top-p. After top scoring bounding boxes are obtained we can use alpha composition to insert the object into the image. An example of such compositing is given in Figure \ref{fig:teaser} where the same steps are applied to place 3 new objects in the original image.

\textbf{Quantitative Results}. Negative log-likelihood (NLL) is a common metric used in previous works for the layout generation task. It is also a good choice for assessing the object insertion task performance since it can be considered as a proxy for class recommendation and bounding box generation accuracy. We outperform current state-of-the-art model \cite{gupta2021layouttransformer} across all datasets in terms of the NLL metric shown in Table \ref{table:nll}.  We also compare LayoutBERT to prior art on COCO in Table \ref{table:nll_coco}. It can be seen that class (Table \ref{table:top-1-class-rec}) and bounding box (Table \ref{table:miou-box-gen}) prediction performance correlates with NLL (Table \ref{table:nll}).

\begin{table*}[ht]
    \centering
    \begin{tabular}{c c c c}
    \toprule
    & LayoutTransformer \cite{gupta2021layouttransformer} & LayoutGPT  & Ours  \\
    \midrule
    COCO \cite{lin2014microsoft} & 2.67 & 2.57 & \textbf{2.32} \\
    PublayNet \cite{zhong2019publaynet} & 1.28  & 1.14 & \textbf{0.63} \\
    Image Layouts  & -  & 2.26 & \textbf{2.08} \\
    Template Layouts & - & 2.19 & \textbf{1.98} \\
    \bottomrule
    \end{tabular}
    \caption{NLL results on COCO, Publaynet, Image Layouts and Template Layouts datasets using 64x64 anchor resolution.}
    \label{table:nll}
\end{table*}

\begin{table}[H]
    \begin{center}
    \begin{tabular}{c c}
    \toprule
    Model & NLL  \\
    \midrule
    LayoutVAE \cite{jyothi2019layoutvae} & 3.29 \\
    ObjGAN \cite{li2019object} & 5.24 \\
    sg2sim \cite{johnson2018image} & 3.4 \\
    LayoutTransformer \cite{gupta2021layouttransformer} & 2.28 \\
    Ours & \textbf{1.91} \\
    \bottomrule
    \end{tabular}
    \end{center}
    \caption{NLL results on COCO using 32x32 anchor resolution. Lower is better.}
    \label{table:nll_coco}
\end{table}

\begin{figure}
    \begin{center}
    \includegraphics[width=\linewidth, scale=2]{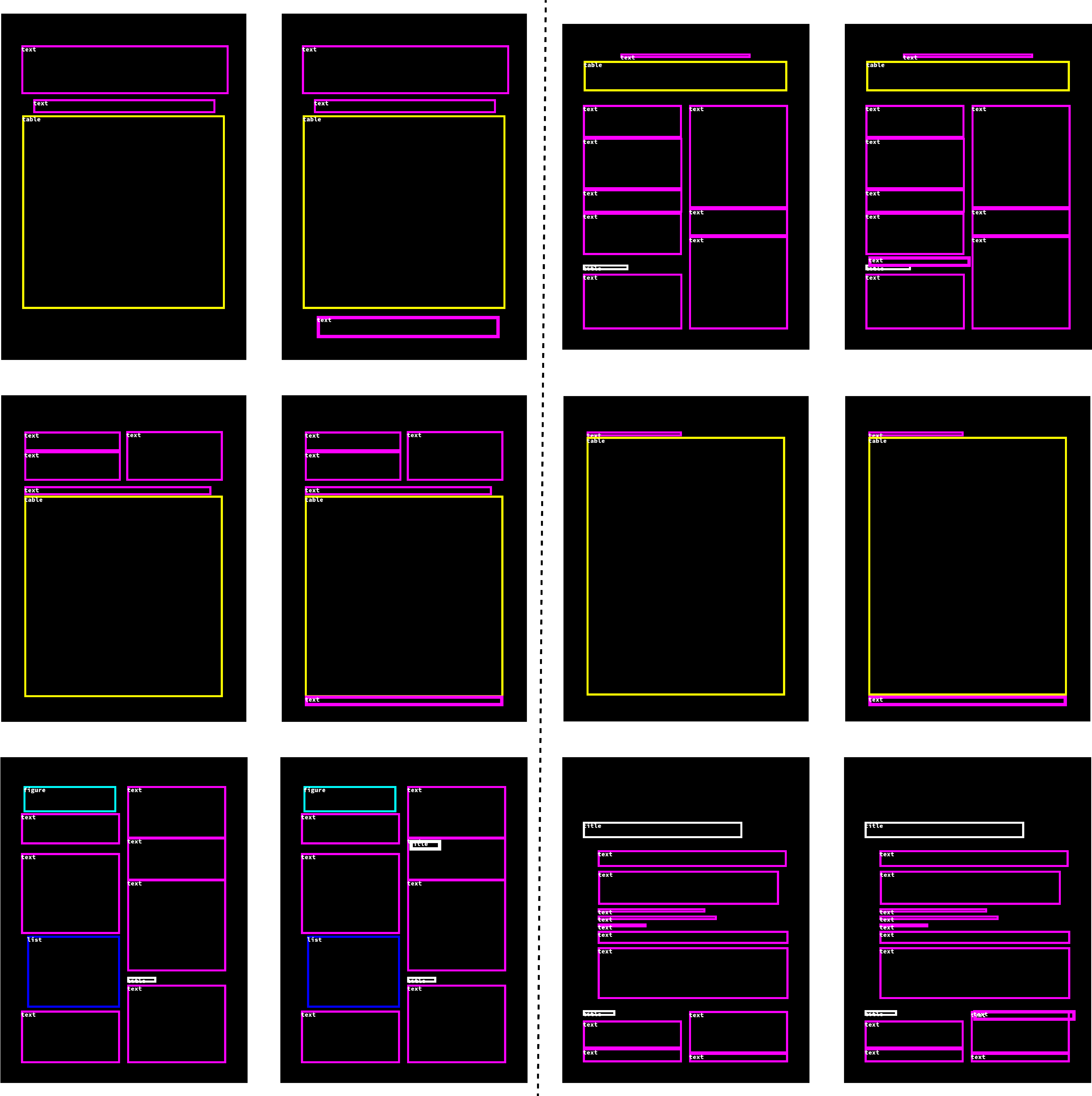}
    \end{center}
    \caption{Side-by-side view of before and after top-1 class bounding box generation on PublayNet validation documents as described in \ref{object-insertion}. Purple: text, yellow: table, white: title, cyan: figure, blue: list. Best viewed in color.}
    \label{fig:publaynet_boxes}
\end{figure}

\begin{figure}
    \begin{center}
    \includegraphics[width=\linewidth, scale=2]{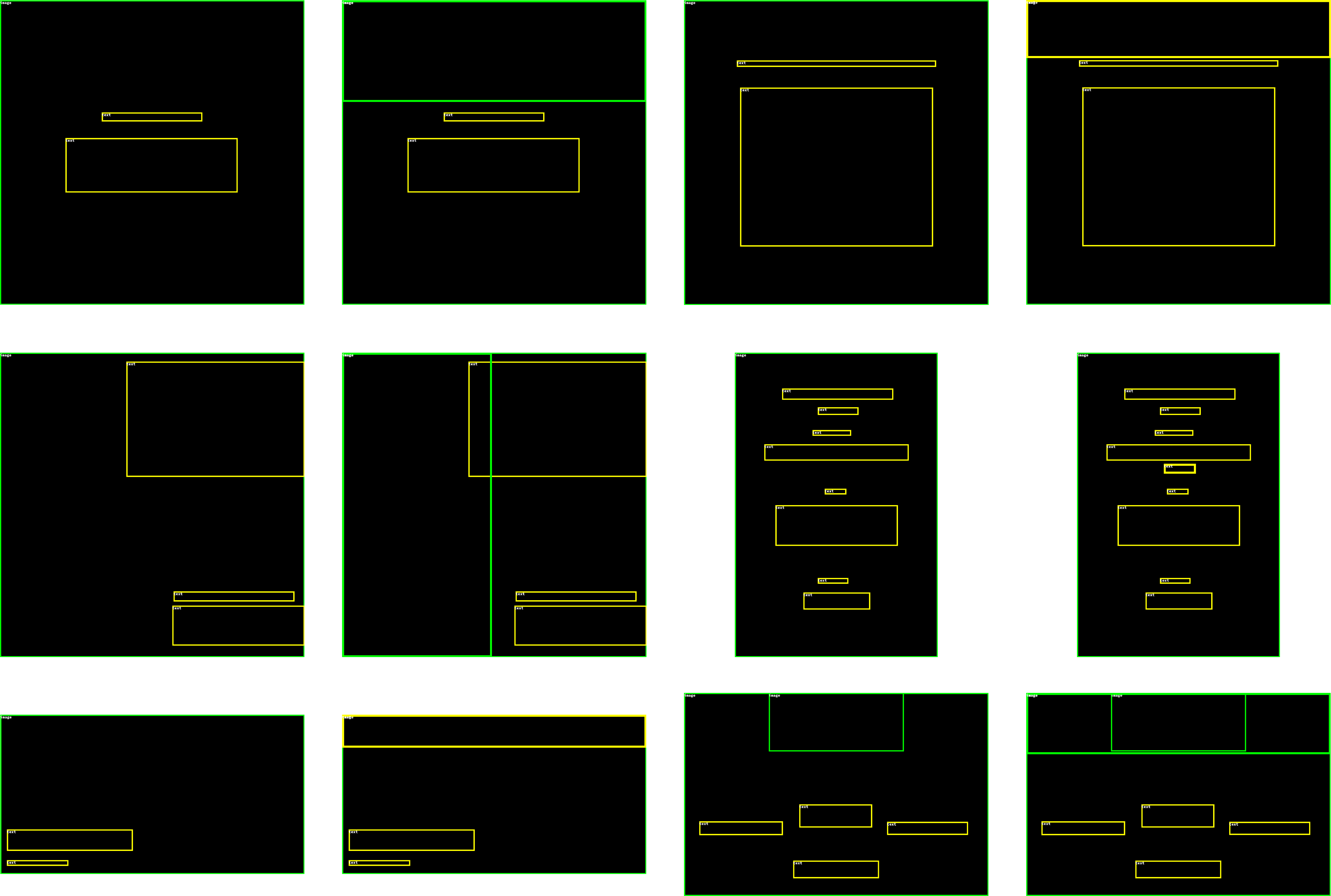}
    \end{center}
    \caption{Side-by-side view of before and after top-1 class bounding box generation on Template Layouts validation templates as described in \ref{object-insertion}. Yellow: text, green: image. Best viewed in color.}
    \label{fig:spark_boxes}
\end{figure}

\begin{figure*}
    \begin{center}
    \includegraphics[width=\linewidth, scale=2]{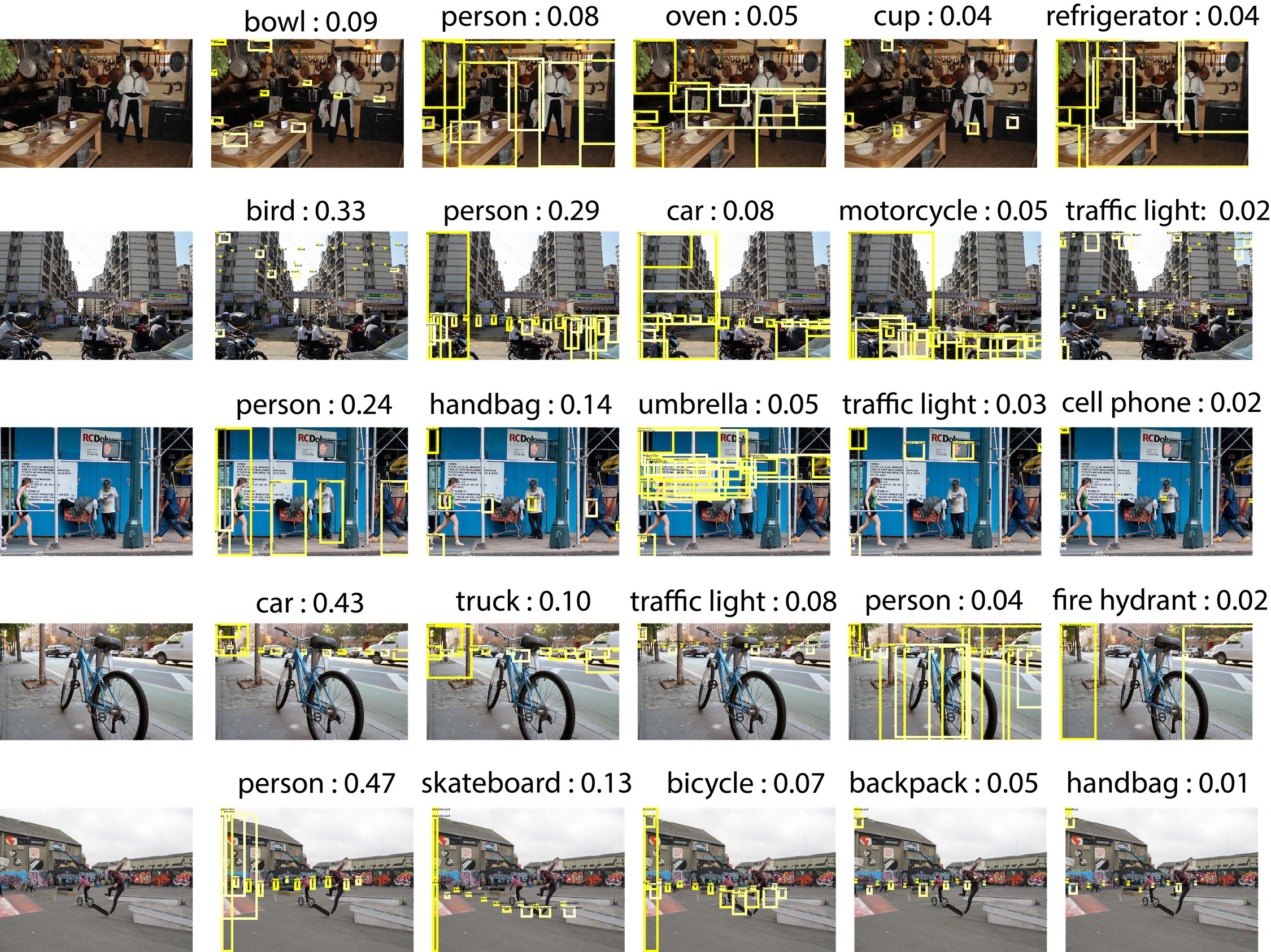}
    \end{center}
    \caption{Class recommendations and bounding box generations conditioned on top-5 classes using COCO validation images as described in \ref{object-insertion}. Predicted classes are sorted from left to right by their associated probability scores. We use k=3 for top-k sampling during bounding box generation. Lighter bounding box color means lower ranking score. Best viewed in color.}
    \label{fig:coco_boxes}
\end{figure*}

\textbf{Qualitative Results}. During qualitative analysis we use the steps described in \ref{object-insertion} and generate composites to be visualized. We pick random samples from the validation set of each dataset, identify top-k classes to be inserted to each sample and then conditionally generate the bounding boxes. In Figure \ref{fig:publaynet_boxes} and Figure \ref{fig:spark_boxes} we insert the most likely bounding box for the top-1 predicted class on PublayNet and Template Layouts datasets respectively, and visualize samples side-by-side before and after the object insertion. In Figure \ref{fig:coco_boxes}, we identify the top-5 classes for each sample from COCO dataset and generate class conditional bounding boxes with top-k sampling to show diverse results.

\subsection{Layout Retrieval}

Each GPT \cite{radford2018improving} feature at a given layer is calculated by attending only to the previous tokens from the previous layer. We tried both the last feature from the last hidden state and the average of all the features from the last hidden state for generating representations for a given layout in the case of the LayoutGPT model. During the qualitative comparison, the average of the last hidden state gave better results, so we use it for extracting representations to be used in the quantitative evaluation. BERT \cite{devlin2018bert} uses bidirectional self-attention so every feature can attend to every other feature from the previous layer. So we use the average of the last hidden state to extract representations in the case of the LayoutBert model.

\textbf{Quantitative Results}. During quantitative analysis, mean average precision at k is used since it is a common choice for retrieval tasks and is widely used for assessing ranked recommendations. None of the datasets used during training were designed for a retrieval-by-layout task in mind, so they do not have relevancy information for a given pair of images, documents or layout templates. We conducted an external assessment job using a data annotation platform which has experienced taskers to manually assess the performance of the models. We ran jobs for the  COCO, PublayNet and Layout Templates datasets and compare the retrieval performance of LayoutGPT and LayoutBERT models. During the COCO evaluation, actual images are shown to taskers, and during the PublayNet and Layout Templates evaluations, rendered layouts with color-coded bounding boxes are shown to taskers. 

We use cosine similarity for retrieving similar layouts and report our results on mAP@5. For a given query the top 5 retrieved layouts are shown for assessment. For each dataset, we use the official validation set for retrieval evaluation. We use 1k random samples as the query set and the remaining as the recall set. Each query is shown to 3 different experienced taskers for final metric calculation which we report in Table \ref{table:map@5}. Final mAP@5 is calculated by taking the weighted average using \emph{tasker trust scores}:

\begin{equation}
\begin{aligned}[b]
mAP@5 = \frac{1}{Q} \sum_{q=1}^{Q=1000}  \sum_{n=1}^{3} AP_{qn} ts_{n}
\end{aligned}
\label{map@5}
\end{equation}

\noindent where $Q$ is the total number of queries, $AP_{qn}$ is the average precision at 5 for query $q$ calculated by tasker $n$, and $ts_{n}$ is the trust score for tasker $n$ normalized to 1. Ease of Job is rated by taskers on a scale of 1 to 5.

\begin{table}[H]
    \begin{center}
    \begin{tabular}{c c c c}
    \toprule
    & LayoutGPT & Ours & Ease of Job \\
    \midrule
    COCO \cite{lin2014microsoft} &  \textbf{0.51} & 0.50 & 3.0/5 \\
    PublayNet \cite{zhong2019publaynet} & 0.29 & \textbf{0.33} & 3.6/5 \\
    Template Lay. & 0.46 & \textbf{0.47} & 2.5/5 \\
    \bottomrule
    \end{tabular}
    \end{center}
    \caption{Retrieval results mAP@5.}
    \label{table:map@5}
\end{table}

\begin{figure}
    \begin{center}
    \includegraphics[width=\linewidth, scale=2]{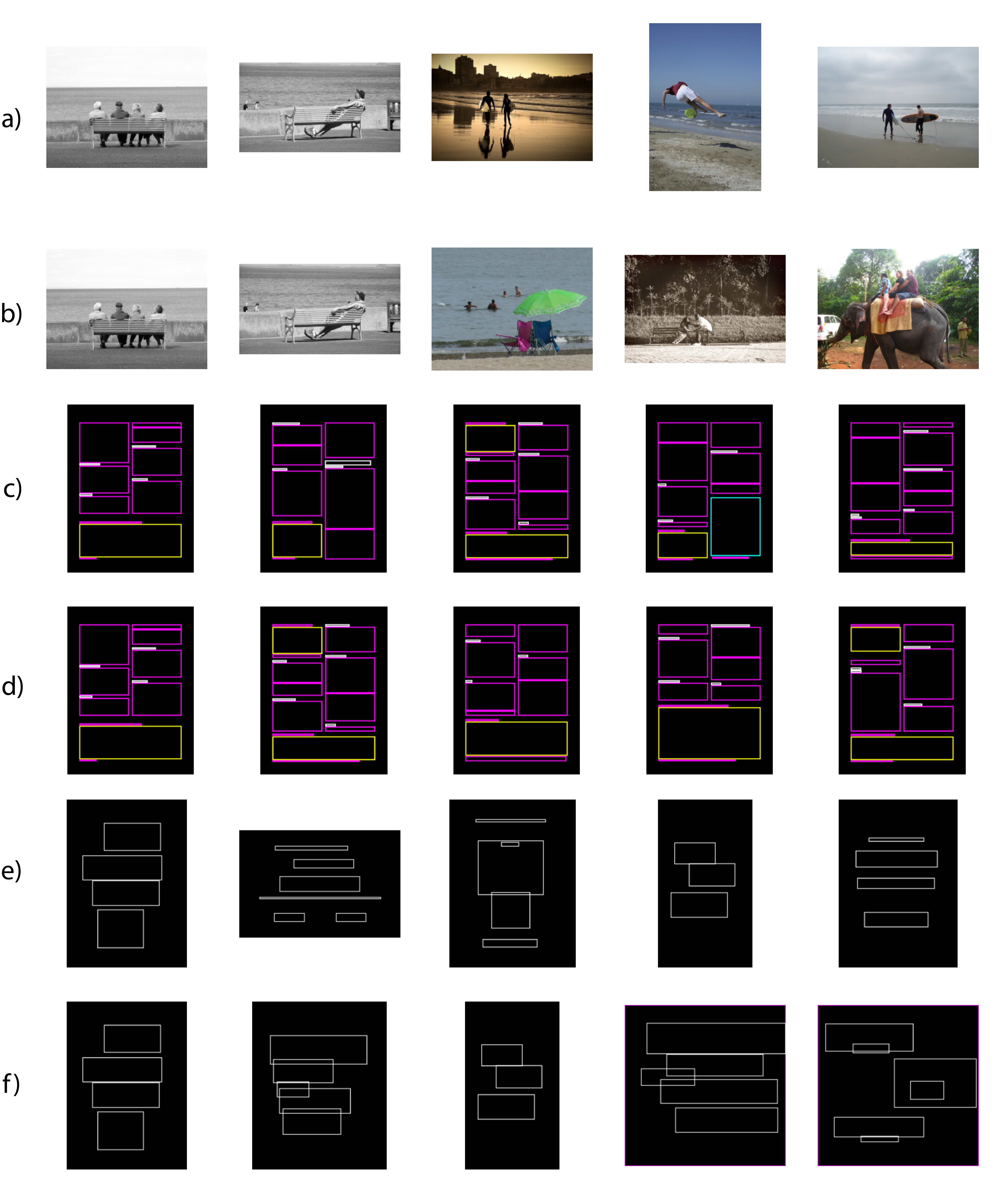}
    \end{center}
    \caption{Retrieval results on COCO, PublayNet and Template Layouts using LayoutGPT and LayoutBERT. Left-most image is the query and the others are top-k neighbors. We show \textbf{(a,b)} LayoutGPT and LayoutBERT on COCO, \textbf{(c,d)} LayoutGPT and LayoutBERT on PublayNet, \textbf{(e,f)} LayoutGPT and LayoutBERT on Template Layouts. Best viewed in color.}
    \label{fig:retrieval}
\end{figure}

\textbf{Qualitative Results}. During qualitative analysis we use the representations extracted from both models, retrieve the top neighbors based on cosine similarity and visualize them side-by-side. Again, we use actual images for COCO and rendered layouts for PublayNet and Layout Templates datasets. An example of retrieval by layout can be seen in Figure \ref{fig:retrieval}.

%------------------------------------------------------------------------
\section{Scaling Studies}

Here we study the effect of the dataset, model and class sample size. We report ablation study results using our large scale Image Layouts dataset.

\textbf{Training Details}. Our small model consists of $d = 256$, $L = 4$, $n_{head} = 4$, and $d_{ff} = 1024$, medium model consists of $d = 512$, $L = 6$, $n_{head} = 8$, and $d_{ff} = 2048$, and large model consists of $d = 768$, $L = 12$, $n_{head} = 12$, and $d_{ff} = 3072$. We also use a dropout of 0.1 at the end of each feed-forward layer for regularization and GELU activation. We use Adam optimizer \cite{kingma2014adam} with decoupled weight decay \cite{loshchilov2017decoupled} with an initial learning rate of $1e{-}3$ using cosine annealing starting after completing 0.75 of training.

\subsection{Effect of Dataset and Model Size}

\begin{figure}
    \begin{center}
    \includegraphics[width=0.8\linewidth, height=0.8\linewidth]{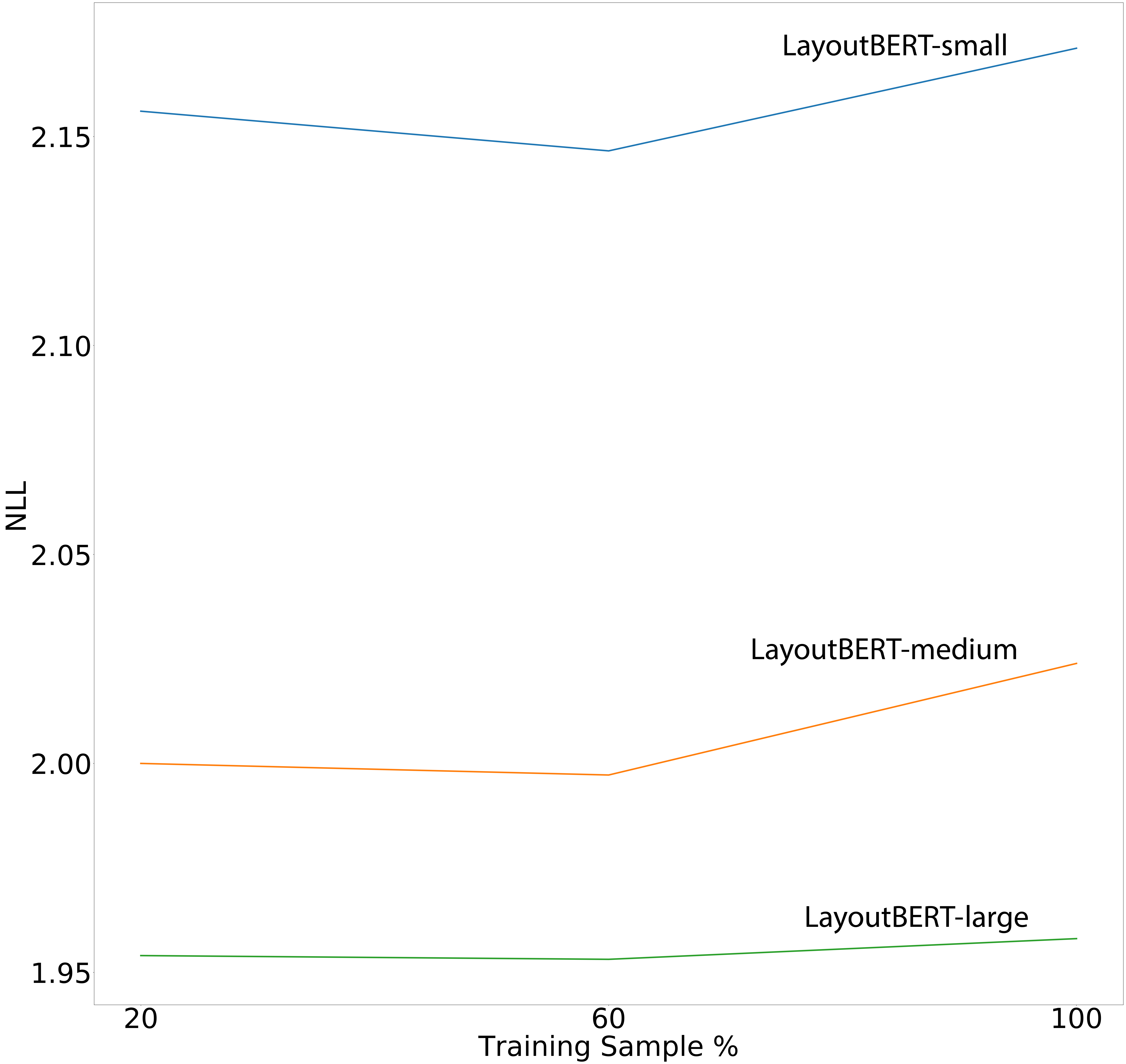}
    \end{center}
    \caption{NLL plots of different model scales and sample sizes on Image Layouts dataset.}
    \label{fig:model_nll}
\end{figure}

We randomly sub-sampled training data in 20\%, 60\%, and 100\% chunks, and 100\% corresponds to 5.8 million layouts from the Image Layouts dataset. Each model is trained with an equal number of forward passes and backward updates, and using the same training schedule for fair comparison. Models with 100\% of the training data are trained for 1.2 epochs, 60\% of the training data are trained for 3 epochs and 20\% of the training data are trained for 6 epochs. We plot results in Figure \ref{fig:model_nll} which shows that small and medium sized models are not able to tolerate an increased number of samples as well as the large sized model. Also, the large model outperforms its smaller counterparts overall. This supports our motivation for creating a large scaled dataset for this task and increasing the model capacity with it. \cite{gupta2021layouttransformer} mentions that they do not observe significant improvements beyond a 6-layer transformer model: However that is not the case with our large scale dataset where LayoutBERT-large shows a 3\% improvement.

\subsection{Effect of Class Sample Size}

\begin{figure}
    \begin{center}
    \includegraphics[width=\linewidth, scale=2]{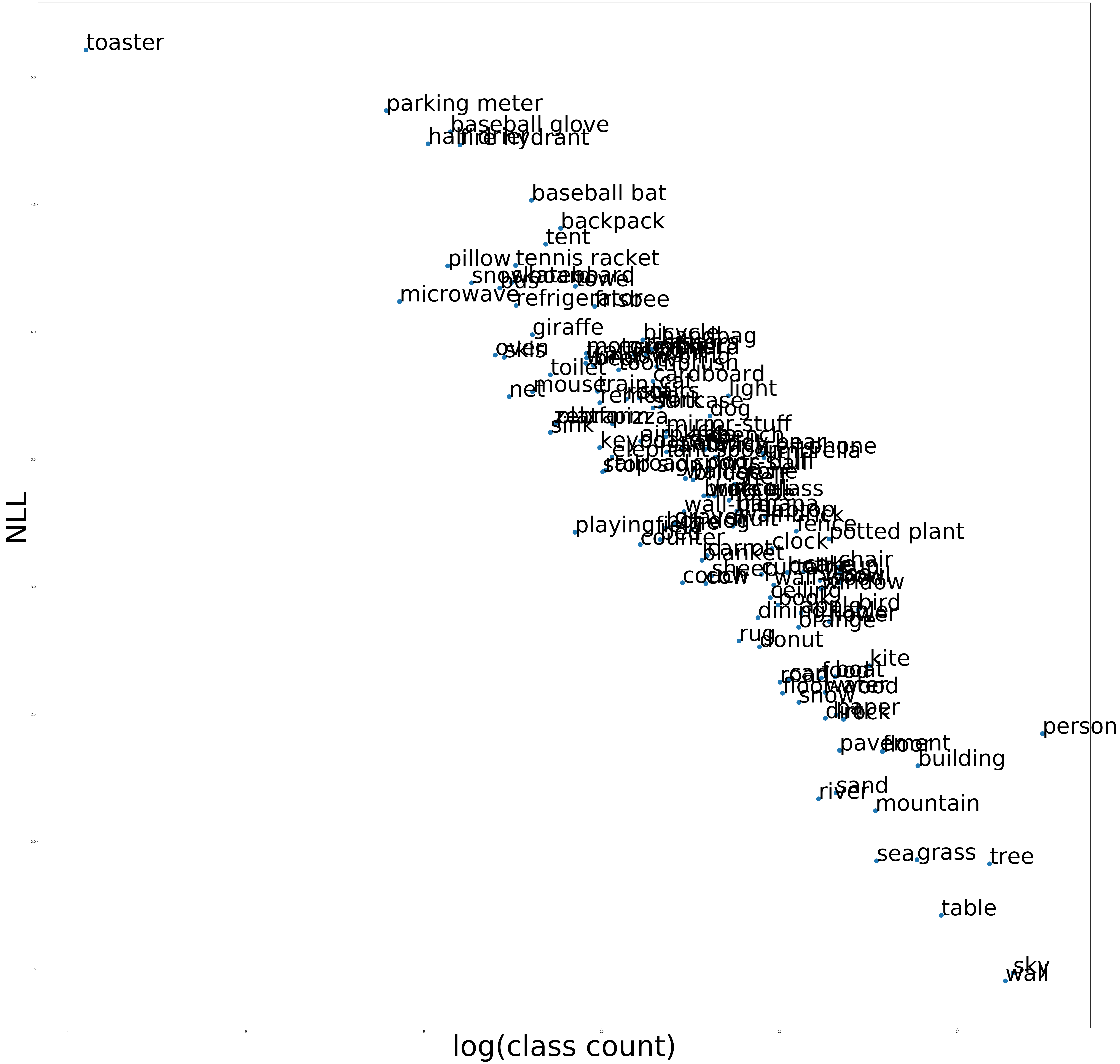}
    \end{center}
    \caption{Class sample size vs. NLL.}
    \label{fig:class_nll}
\end{figure}

We study the performance of each class in our large scale Image Layouts dataset by plotting the NLL per class. There is a positive correlation between class sample size and performance as seen in Figure \ref{fig:class_nll}. Common stuff classes like \emph{sky, wall, sea, tree, grass} and object classes like \emph{person} have low error rates, while rare classes like \emph{toaster} and \emph{parking meter} have much higher error rates. 

%------------------------------------------------------------------------
\section{Conclusion}

Recent works on context-aware object synthesis and layout generation motivated us to find a scalable solution for the object insertion task. Our contributions allow modeling likelihood of hundreds of classes that can be inserted into a given scene by using a novel self-supervised masked language modeling objective and a bidirectional transformer model. Our method pushes the state-of-the-art further on diverse set of data domains like complex scenes, documents and design templates. We also show that deep bidirectional architectures achieve better results as training data size increases.
One drawback of our approach is the limited input representation. In the case of the scene datasets, we lose information by transforming the input signal from the RGB domain to a 2D layout domain by limiting inputs to bounding boxes. In future, we will work on methods which can directly take the raw input data yet leverage the power of transformer models . 

%%%%%%%%% REFERENCES
{\small
\bibliographystyle{ieee_fullname}
\bibliography{main}

\begin{thebibliography}{10}\itemsep=-1pt

\bibitem{arroyo2021variational}
Diego~Martin Arroyo, Janis Postels, and Federico Tombari.
\newblock Variational transformer networks for layout generation.
\newblock In {\em Proceedings of the IEEE/CVF Conference on Computer Vision and
  Pattern Recognition}, pages 13642--13652, 2021.

\bibitem{beltagy2020longformer}
Iz Beltagy, Matthew~E Peters, and Arman Cohan.
\newblock Longformer: The long-document transformer.
\newblock {\em arXiv preprint arXiv:2004.05150}, 2020.

\bibitem{choromanski2020rethinking}
Krzysztof Choromanski, Valerii Likhosherstov, David Dohan, Xingyou Song,
  Andreea Gane, Tamas Sarlos, Peter Hawkins, Jared Davis, Afroz Mohiuddin,
  Lukasz Kaiser, et~al.
\newblock Rethinking attention with performers.
\newblock {\em arXiv preprint arXiv:2009.14794}, 2020.

\bibitem{cordts2016cityscapes}
Marius Cordts, Mohamed Omran, Sebastian Ramos, Timo Rehfeld, Markus Enzweiler,
  Rodrigo Benenson, Uwe Franke, Stefan Roth, and Bernt Schiele.
\newblock The cityscapes dataset for semantic urban scene understanding.
\newblock In {\em Proceedings of the IEEE conference on computer vision and
  pattern recognition}, pages 3213--3223, 2016.

\bibitem{devlin2018bert}
Jacob Devlin, Ming-Wei Chang, Kenton Lee, and Kristina Toutanova.
\newblock Bert: Pre-training of deep bidirectional transformers for language
  understanding.
\newblock {\em arXiv preprint arXiv:1810.04805}, 2018.

\bibitem{girshick2015fast}
Ross Girshick.
\newblock Fast r-cnn.
\newblock In {\em Proceedings of the IEEE international conference on computer
  vision}, pages 1440--1448, 2015.

\bibitem{goodfellow2020generative}
Ian Goodfellow, Jean Pouget-Abadie, Mehdi Mirza, Bing Xu, David Warde-Farley,
  Sherjil Ozair, Aaron Courville, and Yoshua Bengio.
\newblock Generative adversarial networks.
\newblock {\em Communications of the ACM}, 63(11):139--144, 2020.

\bibitem{gupta2021layouttransformer}
Kamal Gupta, Justin Lazarow, Alessandro Achille, Larry~S Davis, Vijay
  Mahadevan, and Abhinav Shrivastava.
\newblock Layouttransformer: Layout generation and completion with
  self-attention.
\newblock In {\em Proceedings of the IEEE/CVF International Conference on
  Computer Vision}, pages 1004--1014, 2021.

\bibitem{jaderberg2015spatial}
Max Jaderberg, Karen Simonyan, Andrew Zisserman, et~al.
\newblock Spatial transformer networks.
\newblock {\em Advances in neural information processing systems},
  28:2017--2025, 2015.

\bibitem{johnson2018image}
Justin Johnson, Agrim Gupta, and Li Fei-Fei.
\newblock Image generation from scene graphs.
\newblock In {\em Proceedings of the IEEE conference on computer vision and
  pattern recognition}, pages 1219--1228, 2018.

\bibitem{jyothi2019layoutvae}
Akash~Abdu Jyothi, Thibaut Durand, Jiawei He, Leonid Sigal, and Greg Mori.
\newblock Layoutvae: Stochastic scene layout generation from a label set.
\newblock In {\em Proceedings of the IEEE/CVF International Conference on
  Computer Vision}, pages 9895--9904, 2019.

\bibitem{kingma2014adam}
Diederik~P Kingma and Jimmy Ba.
\newblock Adam: A method for stochastic optimization.
\newblock {\em arXiv preprint arXiv:1412.6980}, 2014.

\bibitem{kirillov2019panoptic}
Alexander Kirillov, Kaiming He, Ross Girshick, Carsten Rother, and Piotr
  Doll{\'a}r.
\newblock Panoptic segmentation.
\newblock In {\em Proceedings of the IEEE/CVF Conference on Computer Vision and
  Pattern Recognition}, pages 9404--9413, 2019.

\bibitem{kitaev2020reformer}
Nikita Kitaev, {\L}ukasz Kaiser, and Anselm Levskaya.
\newblock Reformer: The efficient transformer.
\newblock {\em arXiv preprint arXiv:2001.04451}, 2020.

\bibitem{lee2018context}
Donghoon Lee, Sifei Liu, Jinwei Gu, Ming-Yu Liu, Ming-Hsuan Yang, and Jan
  Kautz.
\newblock Context-aware synthesis and placement of object instances.
\newblock {\em arXiv preprint arXiv:1812.02350}, 2018.

\bibitem{li2019object}
Wenbo Li, Pengchuan Zhang, Lei Zhang, Qiuyuan Huang, Xiaodong He, Siwei Lyu,
  and Jianfeng Gao.
\newblock Object-driven text-to-image synthesis via adversarial training.
\newblock In {\em Proceedings of the IEEE/CVF Conference on Computer Vision and
  Pattern Recognition}, pages 12174--12182, 2019.

\bibitem{lin2018st}
Chen-Hsuan Lin, Ersin Yumer, Oliver Wang, Eli Shechtman, and Simon Lucey.
\newblock St-gan: Spatial transformer generative adversarial networks for image
  compositing.
\newblock In {\em Proceedings of the IEEE Conference on Computer Vision and
  Pattern Recognition}, pages 9455--9464, 2018.

\bibitem{lin2014microsoft}
Tsung-Yi Lin, Michael Maire, Serge Belongie, James Hays, Pietro Perona, Deva
  Ramanan, Piotr Doll{\'a}r, and C~Lawrence Zitnick.
\newblock Microsoft coco: Common objects in context.
\newblock In {\em European conference on computer vision}, pages 740--755.
  Springer, 2014.

\bibitem{ling2021region}
Jun Ling, Han Xue, Li Song, Rong Xie, and Xiao Gu.
\newblock Region-aware adaptive instance normalization for image harmonization.
\newblock In {\em Proceedings of the IEEE/CVF Conference on Computer Vision and
  Pattern Recognition}, pages 9361--9370, 2021.

\bibitem{liu2019roberta}
Yinhan Liu, Myle Ott, Naman Goyal, Jingfei Du, Mandar Joshi, Danqi Chen, Omer
  Levy, Mike Lewis, Luke Zettlemoyer, and Veselin Stoyanov.
\newblock Roberta: A robustly optimized bert pretraining approach.
\newblock {\em arXiv preprint arXiv:1907.11692}, 2019.

\bibitem{loshchilov2017decoupled}
Ilya Loshchilov and Frank Hutter.
\newblock Decoupled weight decay regularization.
\newblock {\em arXiv preprint arXiv:1711.05101}, 2017.

\bibitem{radford2018improving}
Alec Radford, Karthik Narasimhan, Tim Salimans, and Ilya Sutskever.
\newblock Improving language understanding by generative pre-training.
\newblock 2018.

\bibitem{roberts2020much}
Adam Roberts, Colin Raffel, and Noam Shazeer.
\newblock How much knowledge can you pack into the parameters of a language
  model?
\newblock {\em arXiv preprint arXiv:2002.08910}, 2020.

\bibitem{tsai2017deep}
Yi-Hsuan Tsai, Xiaohui Shen, Zhe Lin, Kalyan Sunkavalli, Xin Lu, and Ming-Hsuan
  Yang.
\newblock Deep image harmonization, 2017.

\bibitem{wang2020linformer}
Sinong Wang, Belinda~Z Li, Madian Khabsa, Han Fang, and Hao Ma.
\newblock Linformer: Self-attention with linear complexity.
\newblock {\em arXiv preprint arXiv:2006.04768}, 2020.

\bibitem{wu2019detectron2}
Yuxin Wu, Alexander Kirillov, Francisco Massa, Wan-Yen Lo, and Ross Girshick.
\newblock Detectron2, 2019.

\bibitem{zhong2019publaynet}
Xu Zhong, Jianbin Tang, and Antonio~Jimeno Yepes.
\newblock Publaynet: largest dataset ever for document layout analysis.
\newblock In {\em 2019 International Conference on Document Analysis and
  Recognition (ICDAR)}, pages 1015--1022. IEEE, 2019.

\end{thebibliography}


\begin{thebibliography}{1}\itemsep=-1pt

\bibitem{kirillov2019panoptic}
Alexander Kirillov, Kaiming He, Ross Girshick, Carsten Rother, and Piotr
  Doll{\'a}r.
\newblock Panoptic segmentation.
\newblock In {\em Proceedings of the IEEE/CVF Conference on Computer Vision and
  Pattern Recognition}, pages 9404--9413, 2019.

\bibitem{lin2014microsoft}
Tsung-Yi Lin, Michael Maire, Serge Belongie, James Hays, Pietro Perona, Deva
  Ramanan, Piotr Doll{\'a}r, and C~Lawrence Zitnick.
\newblock Microsoft coco: Common objects in context.
\newblock In {\em European conference on computer vision}, pages 740--755.
  Springer, 2014.

\bibitem{wu2019detectron2}
Yuxin Wu, Alexander Kirillov, Francisco Massa, Wan-Yen Lo, and Ross Girshick.
\newblock Detectron2, 2019.

\end{thebibliography}
}

\end{document}

% \title{\LaTeX\ Author Guidelines for \confName~Proceedings}
\title{Supplemental Material - LayoutBERT: Masked Language Layout Model for Object Insertion}

% \and
\maketitle

%%%%%%%%% BODY TEXT
\begin{figure}[htp!]
    \begin{center}
    \includegraphics[width=\linewidth,scale=2]{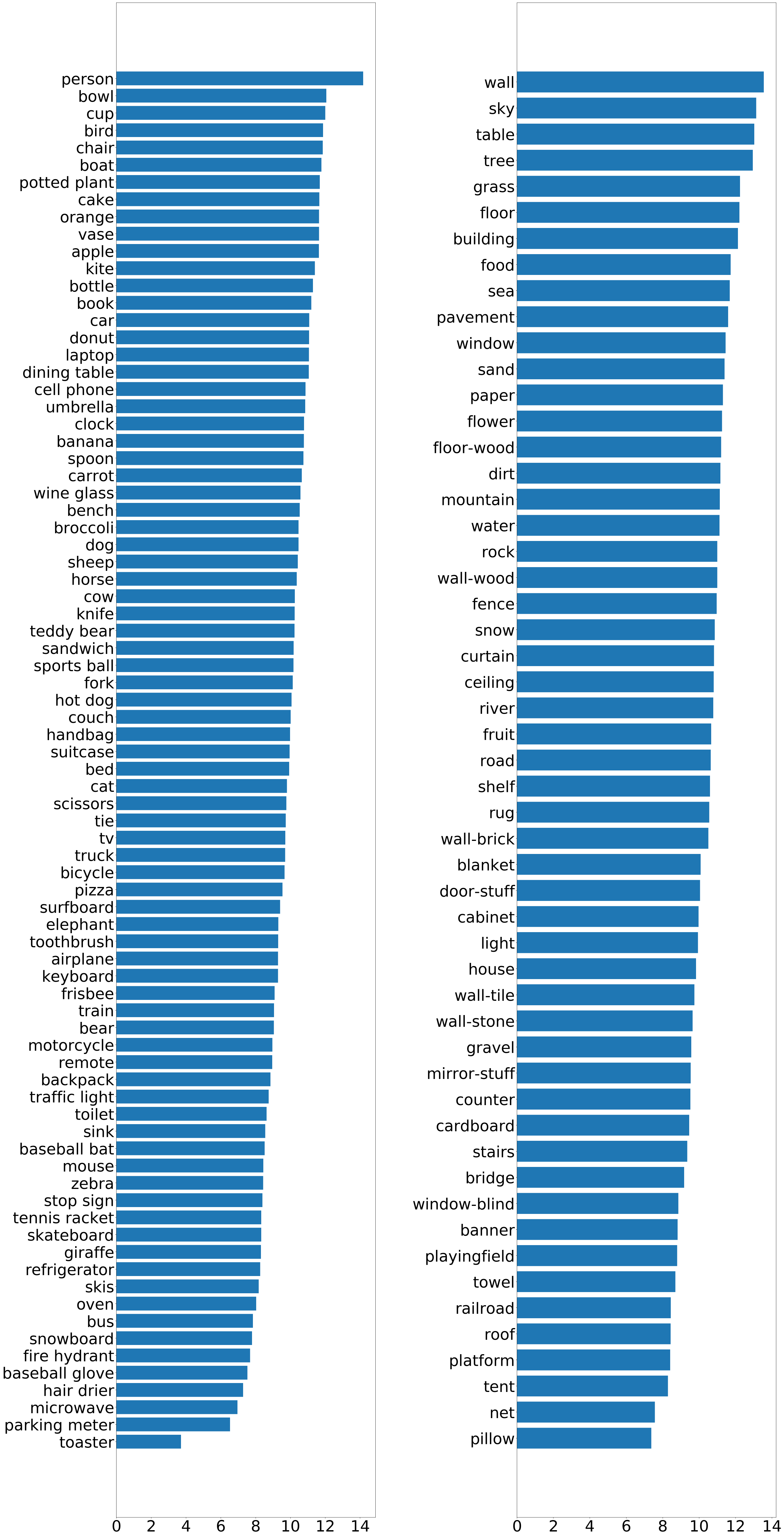}
    \end{center}
    \caption{Thing and stuff class counts in 5.8 million Image Layouts dataset (at log scale).}
    \label{fig:grid}
\end{figure}

\section{Layout Representation}
\label{sec:input}

In this section we will provide more details about layout input representation for training and inference. 

\subsection{Tokenization}

An input sequence consists of \emph{c,x,y,w,h} - class, top-left x coordinate, top-left y coordinate, width and height tokens. Each unique class is tokenized by mapping it to a unique token id starting from 0 to $C - 1$ where $C$ is total number of unique classes in the dataset. Before tokenization we first divide the 2D input into a $NXN$ grid with equal spacing. We call each grid cell an anchor and assign ids by enumerating them from 0 to $N-1$ in both x and y direction. 

For a given bounding box, top-left x and top-left y coordinates can be assigned to the anchor they are in. For example, the chair with the grey bounding box from Figure \ref{fig:grid} has token id assignments of $x-0$ and $y-3$. Using a ruler analogy and the same sizes from the existing grid, width and height tokens are assigned by identifying the minimum number anchors needed to occupy the given width and height. Looking at the same chair example from Figure \ref{fig:grid} we see that width (yellow line) occupies 3 grids and assigned to $w-2$, and height (green line) occupies 4 grids and assigned to $h-3$.

We also have 2 special tokens BOS - beginning of sentence and EOS - end of sentence. 

Once everything taken into account our model has total of $C + 4N +2$ unique tokens, representing classes, x,y,w,h and special tokens.

\begin{figure}[htp!]
    \begin{center}
    \includegraphics[width=\linewidth,scale=2]{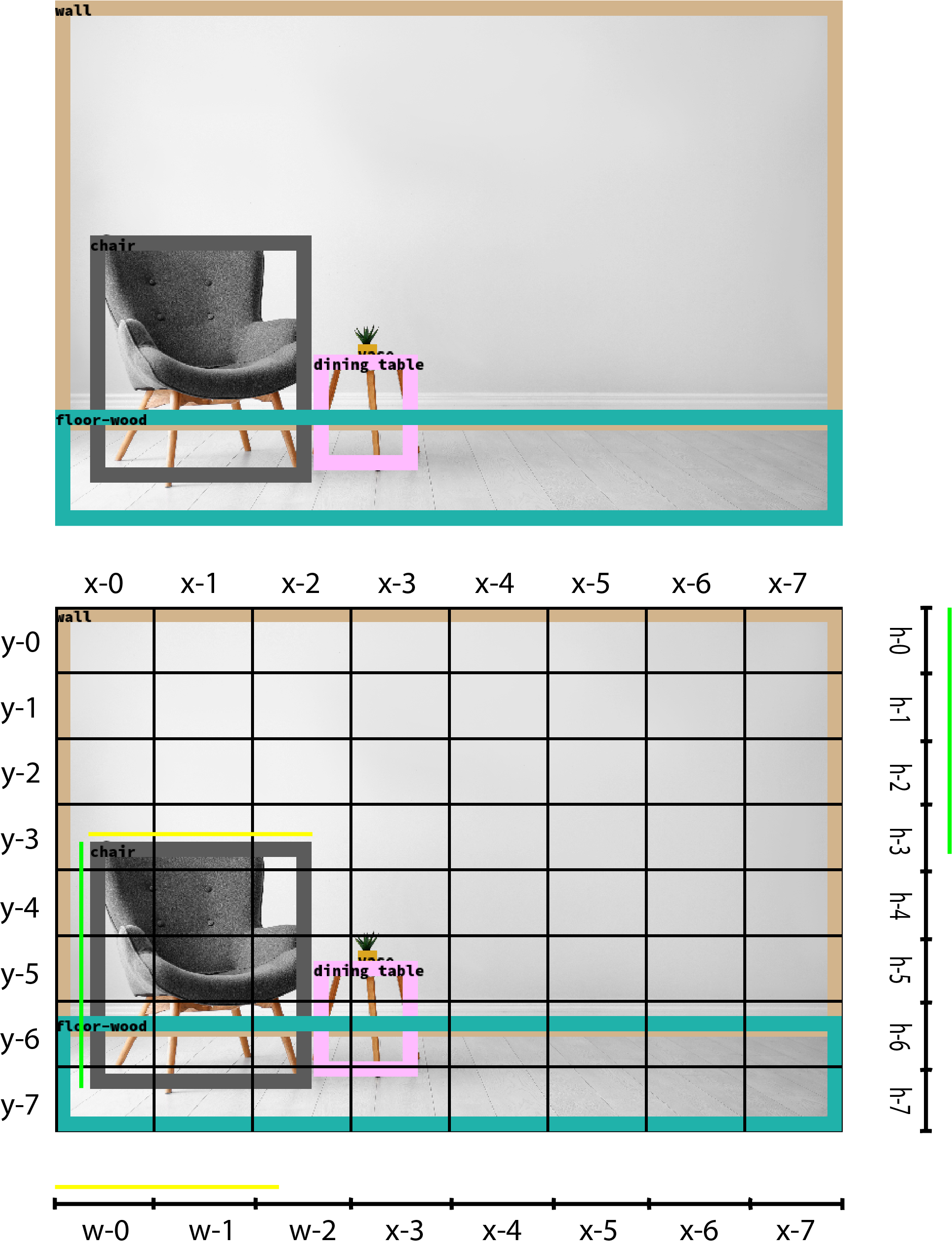}
    \end{center}
    \caption{Grid structure for tokenizing bounding box coordinates. Best viewed in color.}
    \label{fig:grid}
\end{figure}

\section{Image Compositing}
\label{sec:image_compositing}

\begin{figure}[htp!]
    \begin{center}
    \includegraphics[width=\linewidth,scale=2]{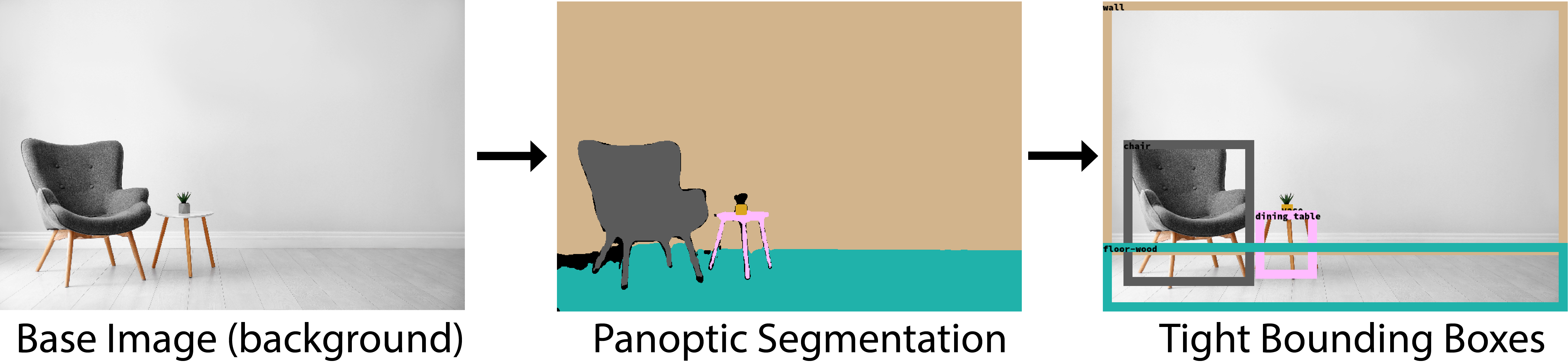}
    \end{center}
    \caption{Input image panoptic segmentation followed by tight bounding box extraction.}
    \label{fig:box_extraction}
\end{figure}

LayoutBERT is trained on object insertion task which is an essential part of image compositing. During inference for image compositing we first feed the base image, the image that new objects will be placed in, to a pretrained panoptic segmentation model \cite{kirillov2019panoptic} using checkpoints from \cite{wu2019detectron2}. It is always possible to train or finetune a new panoptic segmentation model depending on the dataset and the required set of classes for a given use case, but for our demonstration purposes available models pretrained on COCO \cite{lin2014microsoft} suffices. We then convert the predicted panoptic segmentation masks into tight bounding boxes as seen in Figure \ref{fig:box_extraction} and further create the input sequences to be used for either object class recommendations or bounding box generations.

\subsection{Object Class Recommendation}

During object class recommendation we search over all possible positions in the input sequence and extract class probability predictions. Then we take max over all positions for each class prediction to assign the final probability per class.

\begin{figure}[htp!]
    \begin{center}
    \includegraphics[width=\linewidth,scale=2]{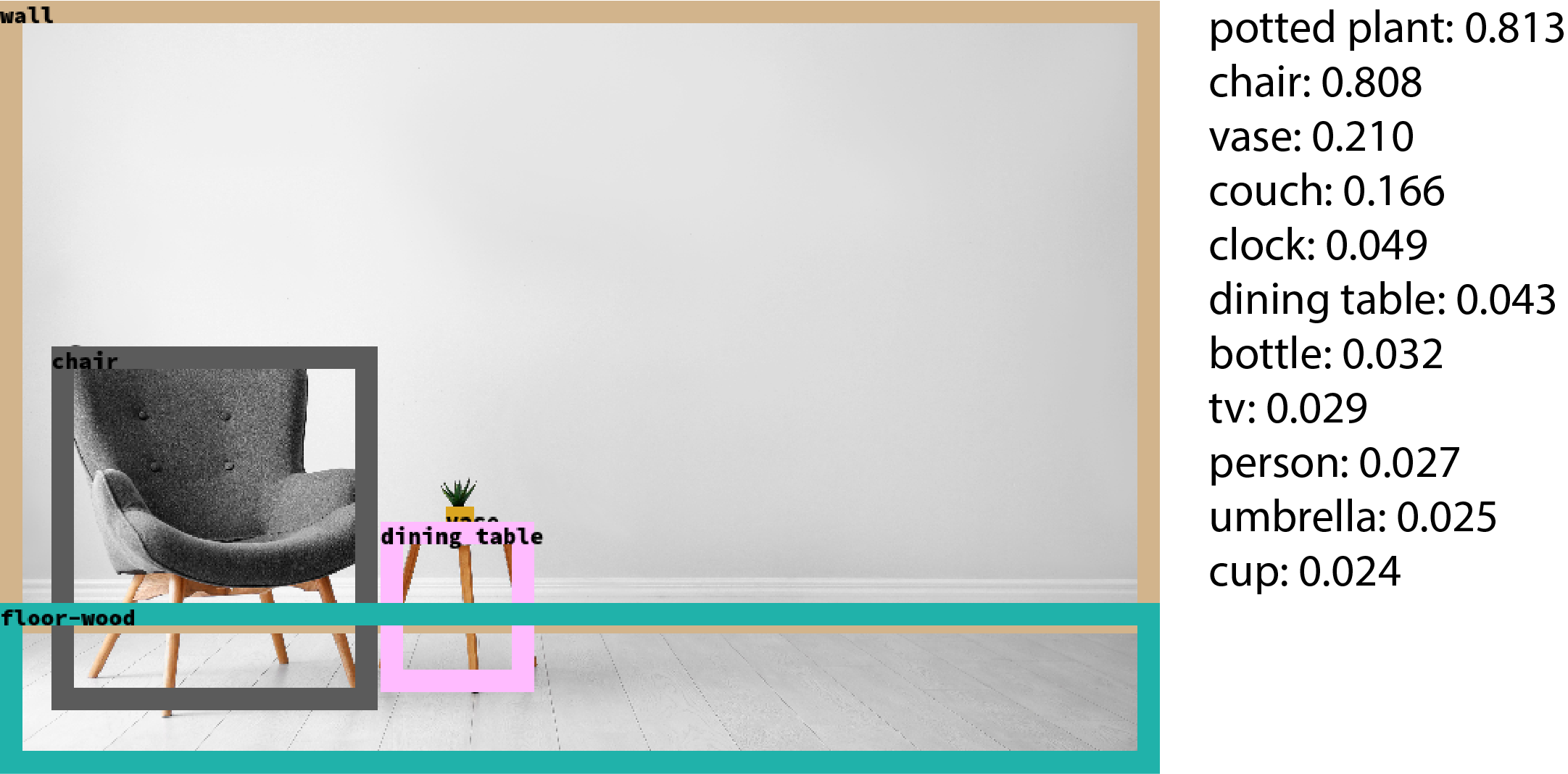}
    \end{center}
    \caption{An example of top class recommendations given the scene layout, class:probability score.}
    \label{fig:class_rec}
\end{figure}

\subsection{Bounding Box Generation}

During class conditional bounding box generation we insert class condition token to every possible position in the input sequence and generate \emph{x,y,w,h} tokens and use top-k sampling for diversity, where k=4 is often a good heuristic in our datasets. Later, we apply non-maximum suppression (NMS) to remove overlapping boxes and finally rank them by using the generated bounding box likelihood score including the class probability at a given sequence position:

\begin{equation}
\begin{aligned}[b]
    p(\theta_{box}) = p(\theta_{c})p(\theta_{x})p(\theta_{y})p(\theta_{w})p(\theta_{h})
\end{aligned}
\label{joint_probability}
\end{equation}

\begin{figure*}
    \begin{center}
    \includegraphics[width=0.75\linewidth]{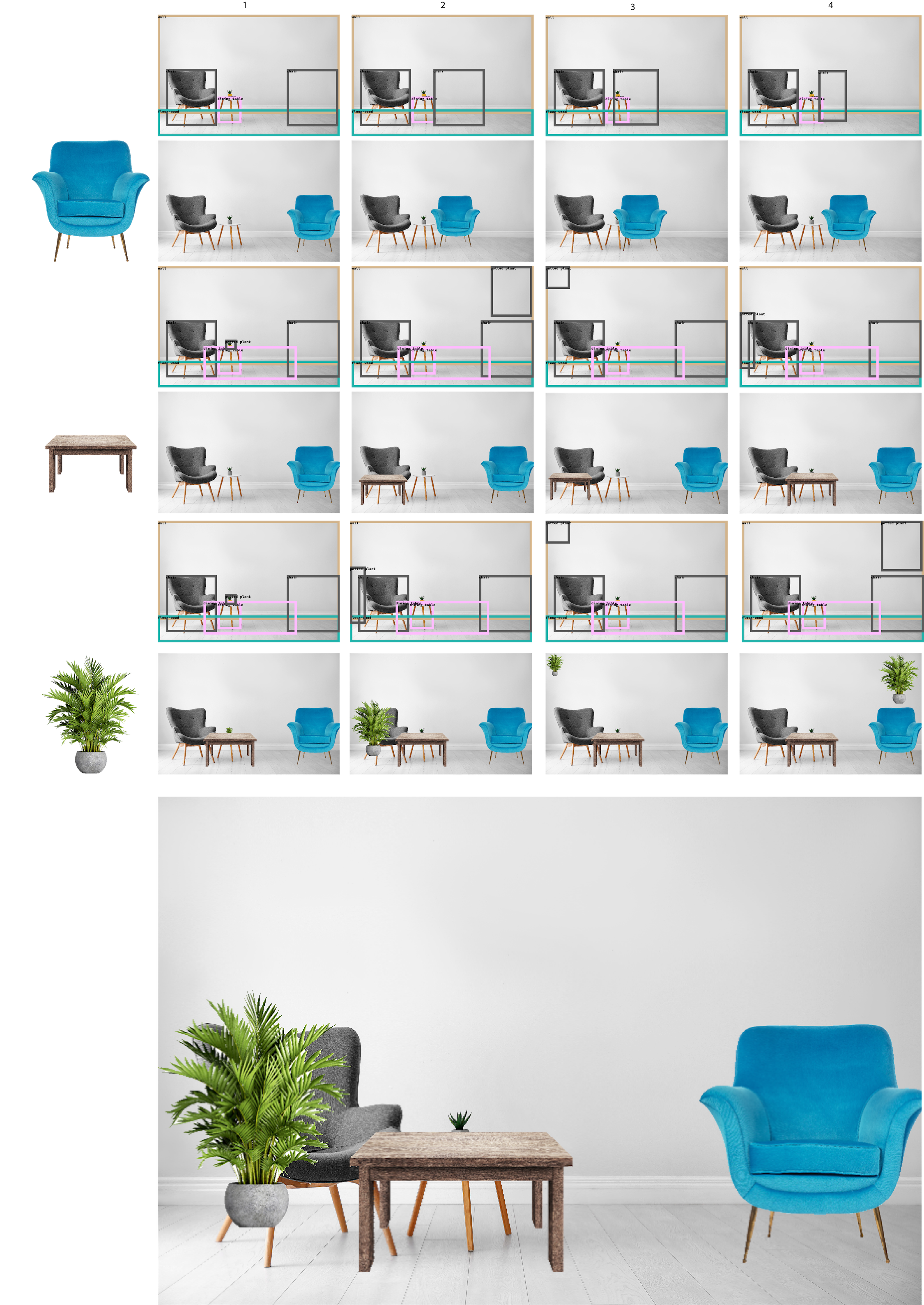}
    \end{center}
    \caption{An example of iterative class conditional bounding box generations and compositing using simple alpha blending. Generated bounding boxes are sorted by their ranking score and displayed from left to right. From top to bottom conditioned object classes are \textbf{\emph{chair, dining table, and potted plant}}. Model trained with Image Layouts dataset is used.}
    \label{fig:compositing_example1}
\end{figure*}

\begin{figure*}
    \begin{center}
    \includegraphics[width=0.75\linewidth]{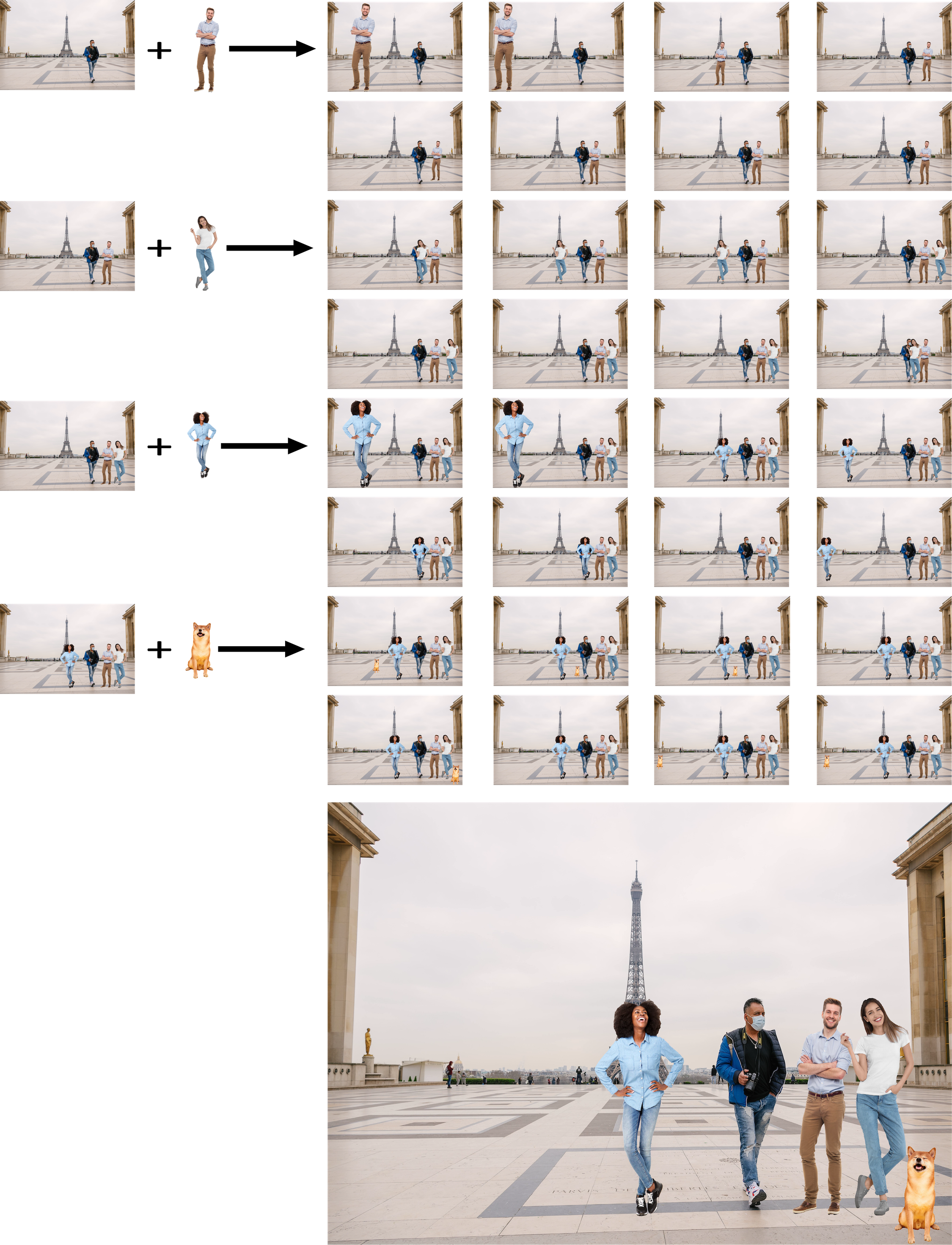}
    \end{center}
    \caption{An example of iterative image compositing. We use model generated bounding boxes conditioned on the class of the object and visualize top image composites on the right. At each row a composite is selected and used as input to the model for generation. From top to bottom conditioned object classes are \textbf{\emph{person, person, person and dog}}. Model trained with Image Layouts dataset is used.}
    \label{fig:compositing_example2}
\end{figure*}

\begin{figure*}
    \begin{center}
    \includegraphics[width=0.75\linewidth]{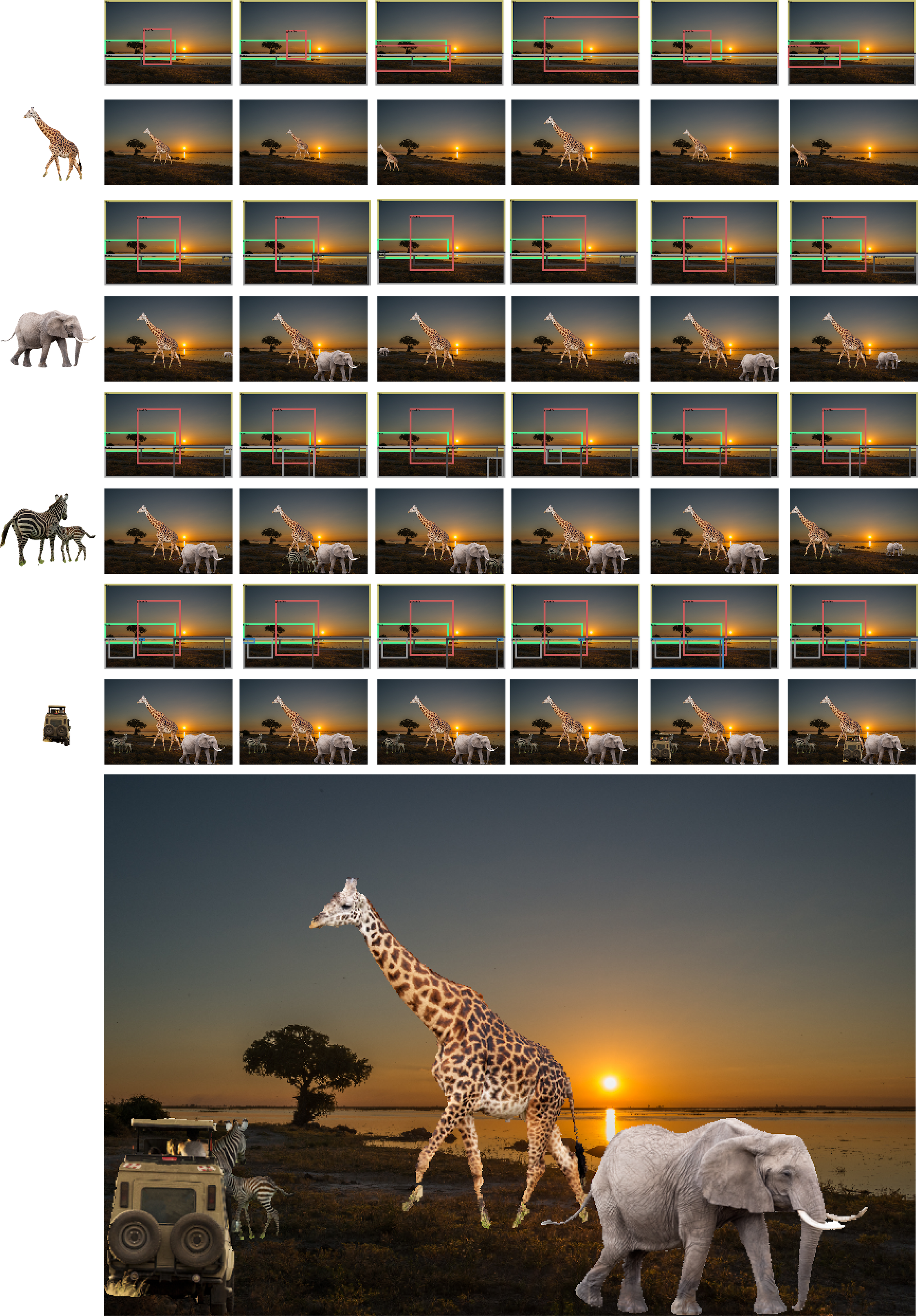}
    \end{center}
    \caption{An example of iterative image compositing. We use model generated bounding boxes conditioned on the class of the object and visualize top image composites on the right. At each row a composite is selected and used as input to the model for generation. From top to bottom conditioned object classes are \textbf{\emph{giraffe, elephant, zebra and car}}. Model trained with COCO dataset is used.}
    \label{fig:compositing_example3}
\end{figure*}

\begin{figure*}
    \begin{center}
    \includegraphics[width=0.9\linewidth]{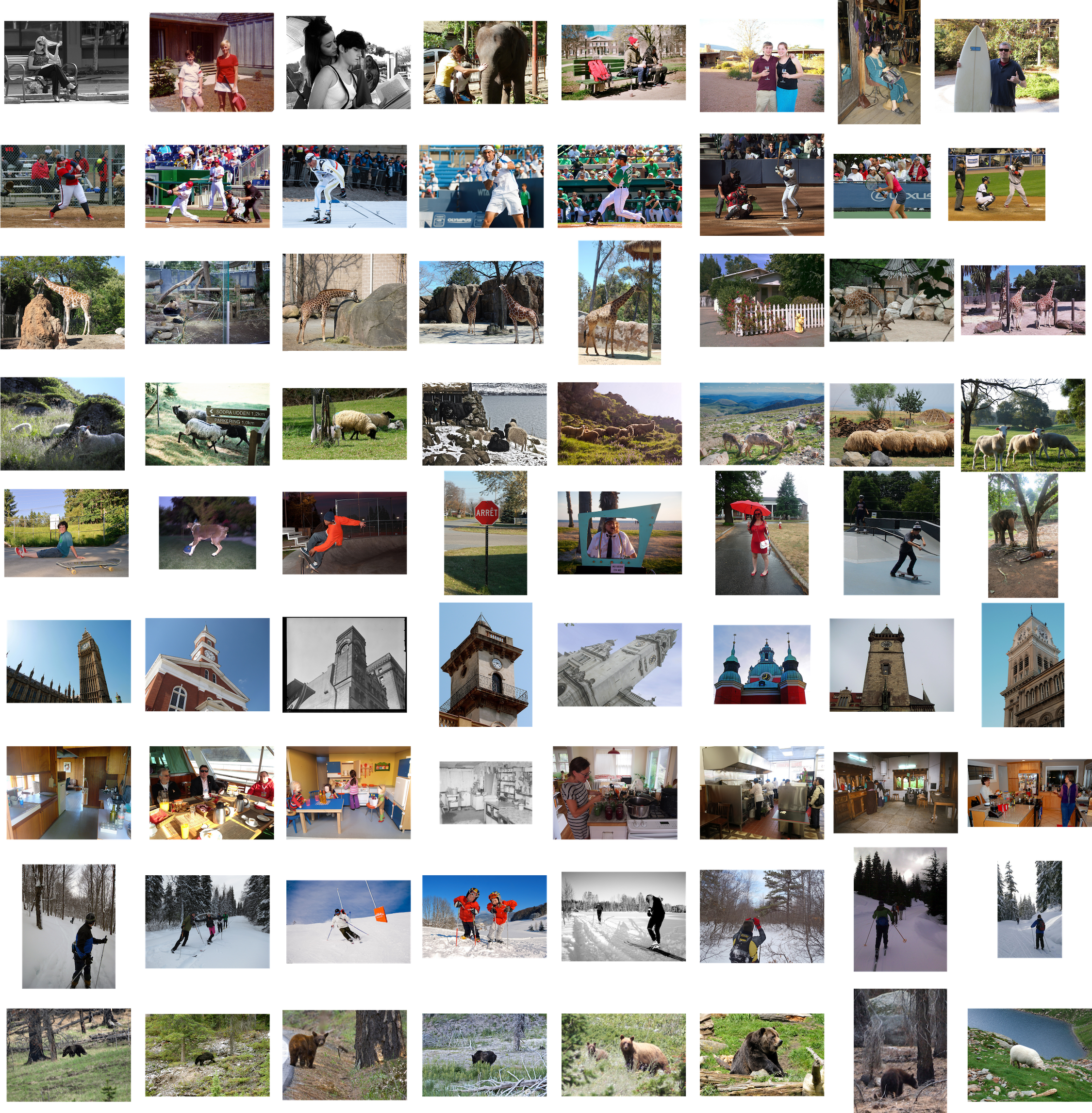}
    \end{center}
    \caption{Retrieval by layout examples. At each row we see the query (left-most image) and the top retrieved images ranked from left to right. Model trained with COCO dataset is used.}
    \label{fig:compositing_example3}
\end{figure*}

%%%%%%%%% REFERENCES
{\small
\bibliographystyle{ieee_fullname}
\bibliography{supplemental}
}